\pdfoutput=1
\documentclass{article}
\usepackage{spconf,amsmath,graphicx}
\usepackage{amsmath}

\title{Inferring a third spatial dimension from 2D histological images}

\name{Maxime W. Lafarge\qquad Josien P.W. Pluim\qquad Koen A.J. Eppenhof\qquad Pim Moeskops\qquad Mitko Veta}

\address{Medical Image Analysis Group, Department of Biomedical Engineering,\\
Eindhoven University of Technology, The Netherlands}
\begin{document}
%\ninept
%
\maketitle
\begin{abstract}
Histological images are obtained by transmitting light through a tissue specimen that has been stained in order to produce contrast.
This process results in 2D images of the specimen that has a three-dimensional structure.
In this paper, we propose a method to infer how the stains are distributed in the direction perpendicular to the surface of the slide for a given 2D image in order to obtain a 3D representation of the tissue.
This inference is achieved by decomposition of the staining concentration maps under constraints that ensure realistic decomposition and reconstruction of the original 2D images.
Our study shows that it is possible to generate realistic 3D images making this method a potential tool for data augmentation when training deep learning models.
\end{abstract}
\section{Introduction}
\label{sec:intro}
In clinical context, pathological diagnosis and prognosis commonly results from the analysis of bright-field microscopy images of histological slides.
These 2D images are obtained by transmitting light through the histological specimens, stained beforehand, in order to attenuate light and produce contrast.
To quantify biomarkers of interest in 2D images, pathologists rely on their experience and knowledge of the 3D context of the objects they observe, when 3D microscopy techniques are not considered.

Taking inspiration from the image formation process of bright-field microscopy, we propose a method to infer a realistic decomposition of hematoxylin and eosin (H\&E) stained histological slides along the axis of their thickness ($z$-axis), resulting in 3D images.
The decomposition of a given histological image is achieved by generating a volume of its underlying stain concentrations,
such that new images obtained by simulating transmitted light along other directions are realistic according to a trained discriminative deep learning model.
%Our method is unsupervised in the way it does not require ground-truth 3D volumes to make inference, but only uses regular 2D images to model how new generated images should appear.

This study is motivated by the recent developments in deep generative models \cite{goodfellow2014generative}, in particular for generating biological microscopy images \cite{osokin2017gans}.
In \cite{gadelha20163d} the authors showed that it is possible to train a generative adversarial network to infer 3D volumes from 2D training images only, without having to rely on 3D training data.
Likewise, our method trains a discriminator from 2D training images only, but can generate 3D volumes that correspond to the decomposition of 2D images, and therefore does not require a generator drawing samples from a latent space.

The proposed algorithm can be seen as generating realistic 3D scenarios for the 2D observed scenes.
As an example of a possible application, the generated 3D volumes can be used for data augmentation as they allow to create new ``views'' of the same data.
Generalization of deep learning models is a known problem in automated histopathology image analysis, and new augmentation methods can help improving generalization \cite{lafarge2017domain}.
The 3D information inferred by our method can also be used for analysis by synthesis strategies \cite{hejrati2014analysis}, to improve histopathology image analysis models, as it is a way to include the prior that processed objects have a 3D structure.

\section{Method}
\label{sec:method}
\begin{figure}[htb]
\begin{minipage}[b]{1.0\linewidth}
  \centering
  \centerline{\includegraphics[width=1.0\linewidth, trim=0pt 560pt 225pt 5pt, clip]{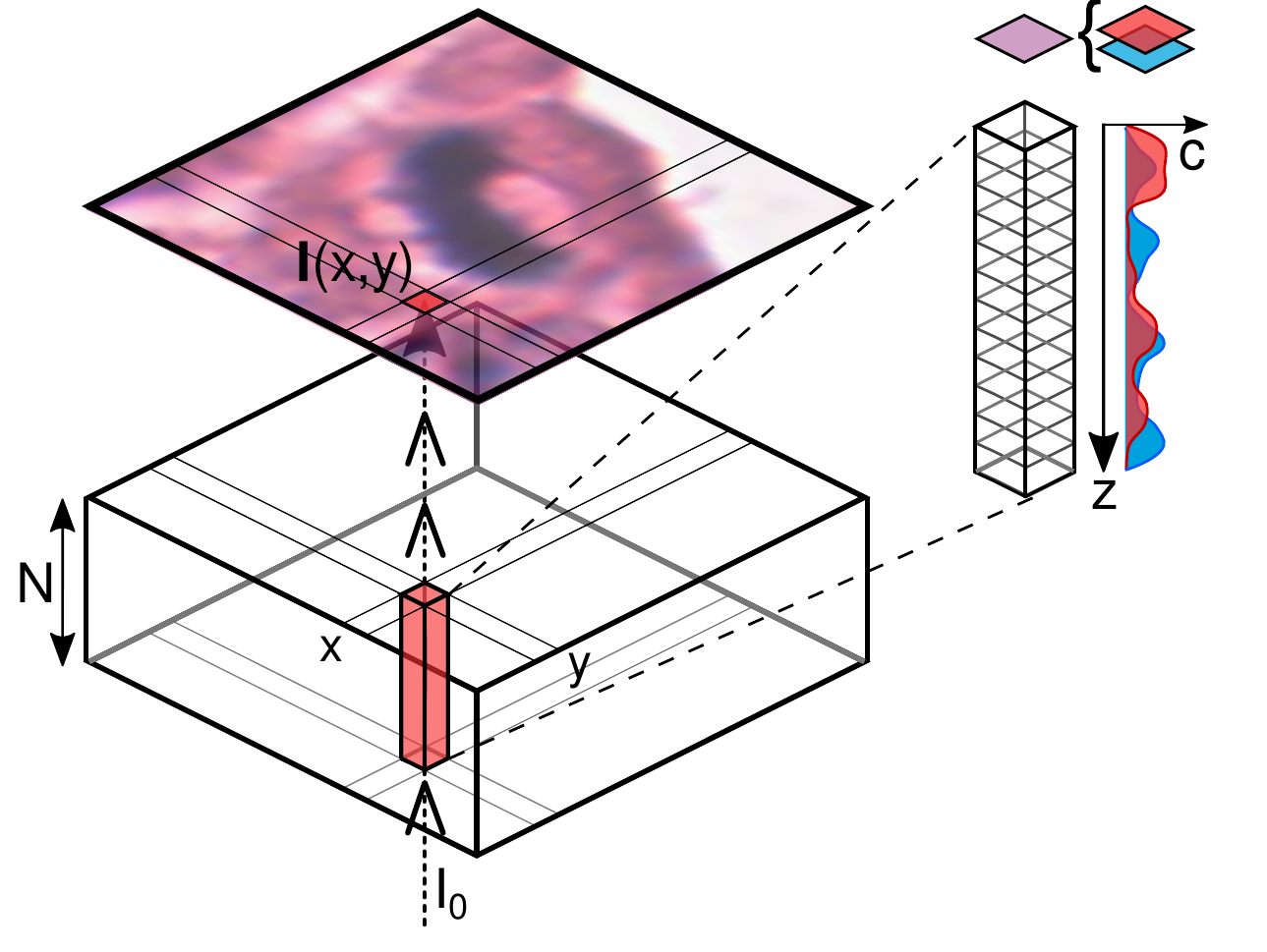}}
%  \vspace{2.0cm}
%  \centerline{(a) Result 1}\medskip
\end{minipage}
\caption{Decomposition of the estimated stain concentration values of a digital slide along the $z$-axis at $(x,y)$.}
\label{fig:decomposition}
\end{figure}

Histological images can be modeled as a set of stain concentrations at every pixel location \cite{ruifrok2001quantification} as illustrated in Fig. \ref{fig:decomposition}.
Thus, our method aims at solving the inverse problem of estimating the volume of stain concentrations that produced the original histological image, for a chosen model of light absorption.
We hypothesize that decomposition in depth is possible since the thickness of the histological specimens is of the order of the image resolution. 
Such a volume is generated under two constraints:
(C1) the reconstruction of the original image must be possible from the estimated volume,
and (C2) new images produced from the volume must be realistic.

\subsection{Model of Stain Concentration Volume}
\label{sec:lightModel}
The RGB pixel intensities can be modeled according to the Beer-Lambert law of light absorption \cite{ruifrok2001quantification},
such that the image intensity at each pixel location $(x,y)$ can be decomposed as 
$I_{c}(x,y) = I_{0} \exp\left(-\mathbf{A}_{c,*} \mathbf{C}(x,y)\right)$
with $c=1, 2, 3$ the color-channel index, $\mathbf{A} \in [0,+\infty]^{3 \times 2} $ the matrix of absorption coefficients specific to the current image,
and $\mathbf{C}(x,y) \in [0,+\infty]^{2}$ the H\&E stain concentrations.
We used the method of \cite{macenko2009method} to achieve unsupervised staining unmixing of the images.

Based on the same model, the stain concentrations can be discretized along the $z$-axis in $N$ parts,
such that $\mathbf{C}(x,y) = \sum_{z=0}^{N-1} \mathbf{\mathcal{C}}(x,y,z)$.

The constraint (C1) can be enforced by reducing the problem to finding the vectors
$\mathbf{V}(x,y,z) \in [0,1]^{2}$, with
$\mathbf{\mathcal{C}}(x,y,z) = \mathbf{C}(x,y) \odot \mathbf{V}(x,y,z)$ and $\sum_{z=0}^{N-1} \mathbf{V}(x,y,z)=[1, 1]^{\top}$
describing how the concentrations $\mathbf{C}(x,y)$ are distributed along the $z$-axis (the operator $\odot$ is the element-wise multiplication).

\subsection{Simulation of Transmitted Light}
\label{sec:simulation}
\begin{figure}[htb]
\begin{minipage}[b]{1.0\linewidth}
  \centering
  \centerline{\includegraphics[width=0.7\linewidth, trim=10pt 260pt 140pt 5pt, clip]{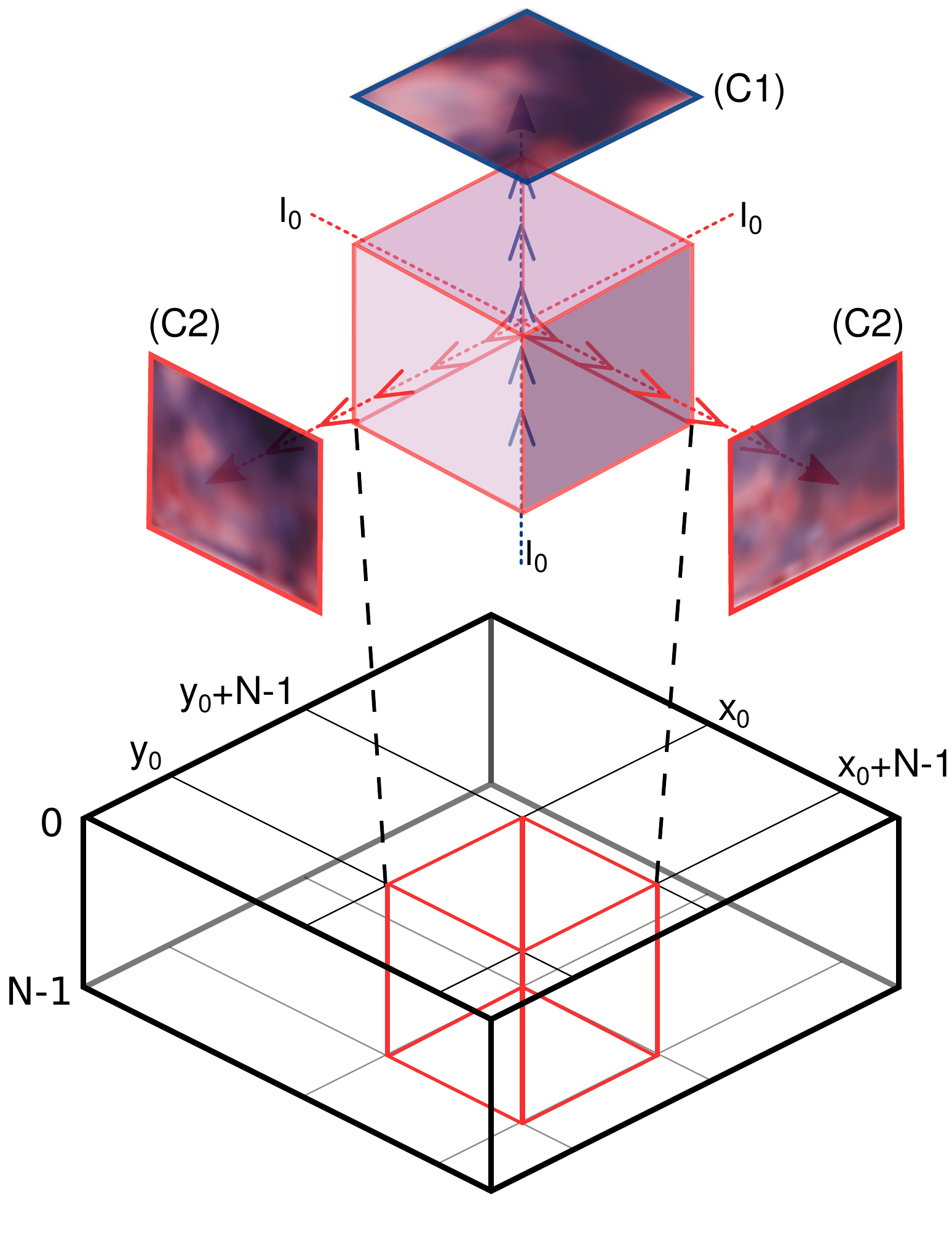}}
%  \vspace{2.0cm}
%  \centerline{.}\medskip
\end{minipage}
\caption{Illustration of an inferred concentration volume block of size $N \times N \times N$ pixels.
(C1) is respected by enforcing the $z$-projection to reconstruct the original image patch.
(C2) requires the $x$/$y$-projections, obtained by simulated transmitted light (red arrows), to be realistic.}
\label{fig:projection}
\end{figure}

For a given volume of concentrations,
new images can be generated by simulating transmitted light from different directions, using the same model of light absorption.
In particular, new projection images $I^{proj}_{x=x_{0}, c}$ and $I^{proj}_{y=y_{0}, c}$
are generated by simulating light transmission along the $x$-axis and $y$-axis through the slices $x \in [x_{0}, x_{0}+N-1]$ and $y \in [y_{0}, y_{0}+N-1]$, as shown in Fig. \ref{fig:projection}.
The pixel intensities of these images are expressed in equation (\ref{eq:projections}) as the sum of stain concentrations in the direction of projection.
\begin{equation}
\label{eq:projections}
\begin{aligned}
I^{proj}_{x=x_{0}, c}(y,z) = I_{0} \exp\left(-\mathbf{A}_{c,*} \sum_{x=x_{0}}^{x_{0}+N-1} \mathbf{\mathcal{C}}(x,y,z)\right) \\
I^{proj}_{y=y_{0}, c}(x,z) = I_{0} \exp\left(-\mathbf{A}_{c,*} \sum_{y=y_{0}}^{y_{0}+N-1} \mathbf{\mathcal{C}}(x,y,z)\right)
\end{aligned}
\end{equation}

$N$ is carefully chosen such that the pixel resolution in the $xz$-slices and $yz$-slices is the 
same as in the original $xy$-plane.

\subsection{Realism Constraint}
\label{sec:realism}
A convolutional neural network can be trained to discriminate ``fake'' generated projection images that result from an underlying unrealistic concentration volume and ``real'' images that are assumed to be the result of realistic volume of concentration distributions.

For a given image patch, 3D volume inference starts from a 4D tensor $\mathbf{V}$ initialized with uniform concentration distributions.
Then, the trained discriminative model (discriminator) can be used to update $\mathbf{V}$ by gradient descent,
so that the generated projections of the updated volume appear slightly more realistic.
The gradient of the loss of the discriminator with respect of the input is computed via back-propagation.

\begin{figure*}[htb]
\begin{minipage}[b]{1.0\linewidth}
  \centering
  \centerline{\includegraphics[width=0.85\linewidth, trim=10pt 585pt 5pt 10pt, clip]{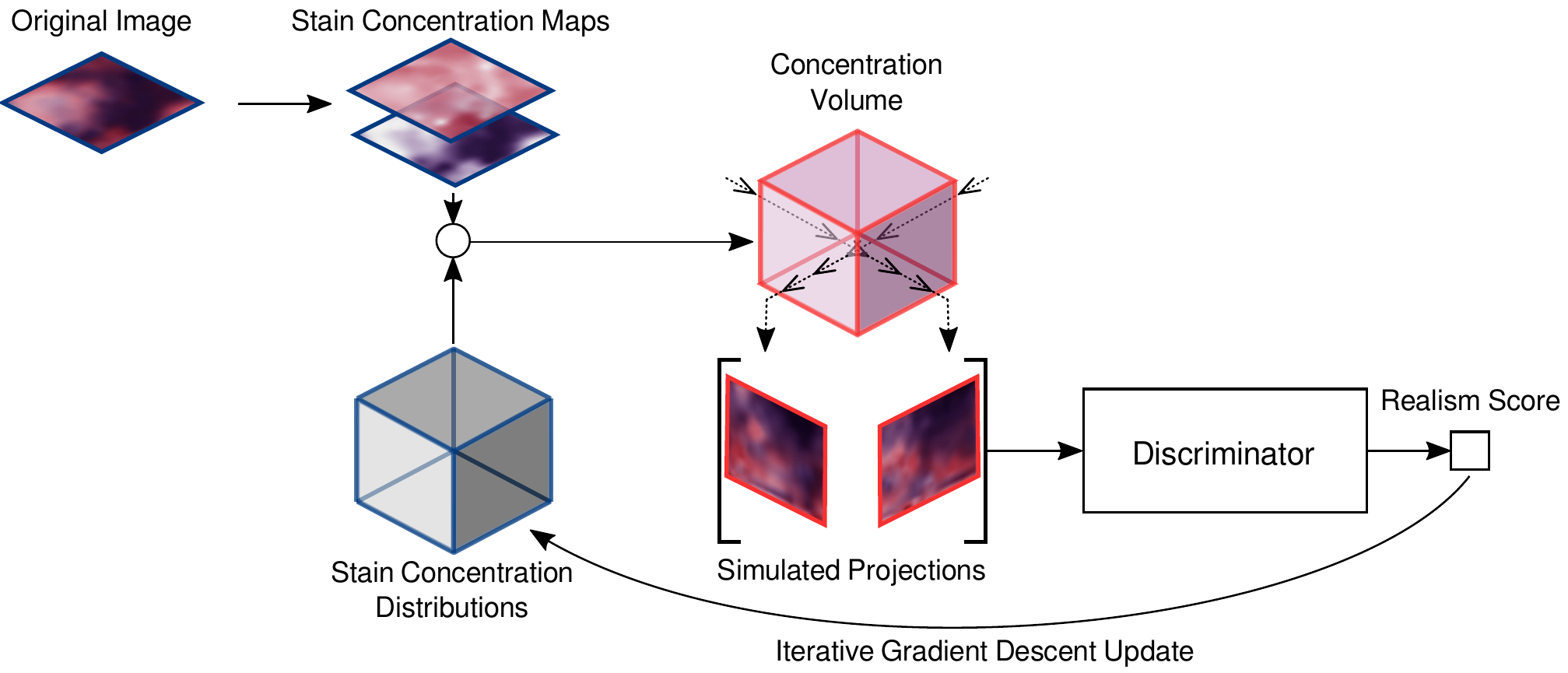}}
\end{minipage}
\caption{Iterative process of generating a volume of concentrations constrained by an original image.
The stain concentration volume is updated by gradient descent in order to produce projection images that ``fool'' the fixed trained discriminator.}
\label{fig:generativeProcess}
\end{figure*}

This update process (Fig. \ref{fig:generativeProcess}) is iterated until convergence:
when the discriminator classifies the generated projections as realistic with small error.

This image generation approach via optimization of the loss function of a neural network is similar to the methods developed in \cite{gatys2015texture,gatys2016image}, and plays a role comparable to the generator of standard generative adversarial networks \cite{goodfellow2014generative} in the way how generated images are used as input to a discriminator. 

\subsection{Discriminator Training}
\label{sec:discriminator}
The discriminator is trained using two sets:
a set of random real image patches $S_{real}$, and a set of adversarial examples $S_{adv}$ that are generated during training,
using the projections computed with (\ref{eq:projections}), from previous states of the trained model.

The training procedure alternates between two steps.
First, the current state of the model is used to infer volumes from real images via gradient update using the process presented in Sect. \ref{sec:simulation}, and $x$/$y$-axis projections produced from this volume are added to  $S_{adv}$.
Secondly, a batch of image patches balanced between samples of $S_{real}$ and $S_{adv}$ is used to train the discriminator.
Images in $S_{adv}$ are sampled according to their misclassification probability such that the model learns from the ``fake'' generated images that are the most realistic and that are more challenging to classify.

%\vfill
%\pagebreak
\section{Experiments and Results}
\label{sec:experiments}

\subsection{Dataset}
We used the high power field images of H\&E stained slides of the public AMIDA13 dataset \cite{veta2015assessment} for the experiments.
232 images of size $2000 \times 2000$ pixels from 8 different breast cancer cases were used for training and the remaining images were used to generate test examples.

\subsection{Discriminator Architecture and Training Procedure}
We implemented the discriminator that can classify input patches as ``fake'' or realistic as a 6-layer convolutional neural network.
The network takes $24\times24$ image patches transformed to H\&E concentration maps as input.
% and was trained using batches of size 64.
Kernels of size $3\times3$, batch normalization, average-pooling and leaky ReLU non-linearities were used throughout.
The network was trained by minimizing the cross-entropy loss using the Adam optimizer.
%Training was stopped after convergence of the generated projections on an independent validation set.

\subsection{Generative Process and Extension to Large Images}
We set the $z$-axis discretization to $24$ pixels as we considered $6$ micrometers as the maximum thickness of a tissue slice,
in which case the $z$-axis pixel resolution of the inferred volumes can be the same as in the $xy$-plane ($0.25$ micrometers).
%During the generative process, the volumes were updated by stochastic gradient descent.

The discriminator, as such, can only infer volumes from images of size $24\times24$.
To overcome this limitation, volumes of larger images can be inferred by optimizing overlapping $24\times24\times24$ sub-volumes in parallel.
This solution was used to produce stain concentration volumes from $64\times64$ images.

\begin{figure*}[ht!]
\centering
\begin{minipage}{0.9\linewidth}
  \centering
  \centerline{\includegraphics[width=1.0\linewidth, trim=15pt 90pt 100pt 5pt, clip]{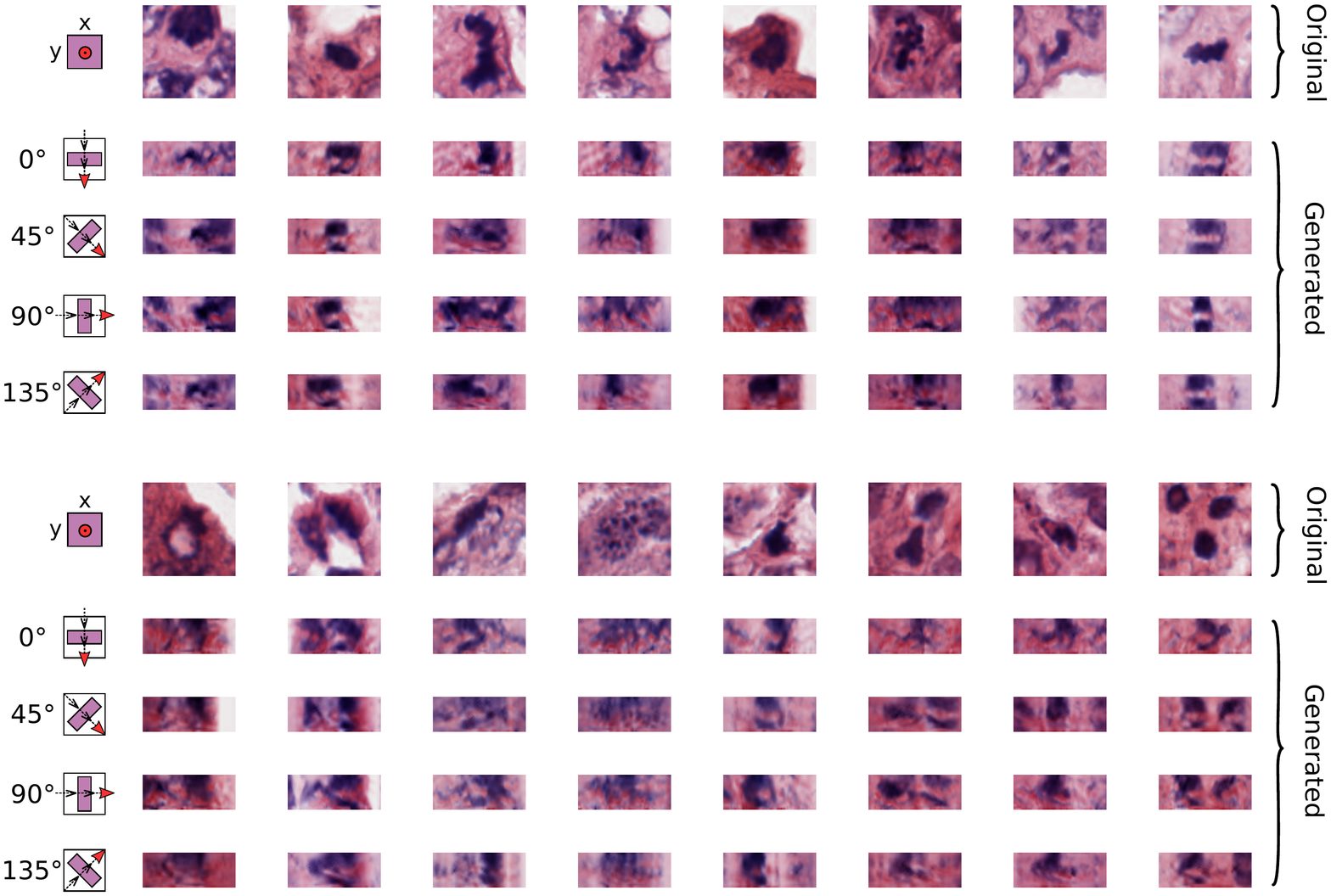}}
%  \vspace{2.0cm}
%  \centerline{(a) Result 1}\medskip
\end{minipage}
\caption{Examples of projection images from generated volumes of stain concentration.
The first row of each block shows the real image patches the volumes were inferred from.
The other rows show the projections obtained by simulating light transmission in different oriented slices as indicated in the left column.
The top block shows results on mitotic figures that were annotated by expert pathologists, and the bottom block includes non-mitotic figures only. }
\label{fig:results}
\end{figure*}

The generated projections presented in Fig. \ref{fig:results} indicate that the optimization process is able to distribute the stain concentrations of unseen images across the $z$-axis, and is able to create new tissue structures that are realistic for the trained discriminator.

\section{Discussion and Conclusions}
\label{sec:discussion}
%We trained a discriminative model, that combined to a 3D model of decomposition of regular 2D histological images, allows to infer 3D images.
%We combined a 3D model of decomposition of regular 2D histological images to trained a discriminative model in %order to infer 3D images.
We proposed a method for inferring the 3D structure of 2D histological images.
The method showed good qualitative performance when applied to an image dataset of mitoses and non-mitosis objects extracted from breast cancer histology slides.
%Our method was applied to infer stain concentration volumes of unseen images, and we interpret the new projection images as having good qualitative appearances, given that direct modalities to produce such images from digital slides do not exist.
Although the volumes generated by our method cannot be considered as representing the actual tissue structure,
the generated projections can still be considered as a likely scenario and thus used as a data augmentation tool.

In addition to being driven by the image formation process of bright-field microscopy, our method has the property that the generated images are directly produced from the available data,
the same way transformation-based augmentation methods work.
In contrast, generators drawing inputs from a latent space, such as generative adversarial networks, do not have this property.
% as the generated images are alternative realistic views of already existing objects.

Directions of future work include, further research to assess the realism quality of the generated images, and application of the generated 3D representation for data augmentation.
%, and training a generative model to directly infer a realistic volume without having to rely on the iterative process we used.

\bibliographystyle{IEEEbib}
\bibliography{refs}

\begin{thebibliography}{10}

\bibitem{goodfellow2014generative}
I~Goodfellow, J~Pouget-Abadie, M~Mirza, B~Xu, D~Warde-Farley, S~Ozair,
  A~Courville, and Y~Bengio,
\newblock ``Generative adversarial nets,''
\newblock in {\em NIPS}, 2014, pp. 2672--2680.

\bibitem{osokin2017gans}
A~Osokin, A~Chessel, RE~Carazo Salas, and F~Vaggi,
\newblock ``{GAN}s for biological image synthesis,''
\newblock in {\em ICCV}, 2017.

\bibitem{gadelha20163d}
M~Gadelha, S~Maji, and R~Wang,
\newblock ``{3D} shape induction from {2D} views of multiple objects,''
\newblock {\em arXiv:1612.05872}, 2016.

\bibitem{lafarge2017domain}
MW~Lafarge, JPW Pluim, KAJ Eppenhof, P~Moeskops, and M~Veta,
\newblock ``Domain-adversarial neural networks to address the appearance
  variability of histopathology images,''
\newblock in {\em MICCAI-DLMIA}, 2017, pp. 83--91.

\bibitem{hejrati2014analysis}
M~Hejrati and D~Ramanan,
\newblock ``Analysis by synthesis: {3D} object recognition by object
  reconstruction,''
\newblock in {\em IEEE CVPR}, 2014, pp. 2449--2456.

\bibitem{ruifrok2001quantification}
AC~Ruifrok, DA~Johnston, et~al.,
\newblock ``Quantification of histochemical staining by color deconvolution,''
\newblock {\em Anal. Quant. Cytol.}, vol. 23, no. 4, pp. 291--299, 2001.

\bibitem{macenko2009method}
M~Macenko, M~Niethammer, JS~Marron, D~Borland, JT~Woosley, X~Guan, C~Schmitt,
  and NE~Thomas,
\newblock ``A method for normalizing histology slides for quantitative
  analysis,''
\newblock in {\em IEEE ISBI}, 2009, pp. 1107--1110.

\bibitem{gatys2015texture}
L~Gatys, AS~Ecker, and M~Bethge,
\newblock ``Texture synthesis using convolutional neural networks,''
\newblock in {\em NIPS}, 2015, pp. 262--270.

\bibitem{gatys2016image}
L~Gatys, AS~Ecker, and M~Bethge,
\newblock ``Image style transfer using convolutional neural networks,''
\newblock in {\em IEEE CVPR}, 2016, pp. 2414--2423.

\bibitem{veta2015assessment}
M~Veta, PJ~{van}~Diest, SM~Willems, H~Wang, A~Madabhushi, A~Cruz-Roa,
  F~Gonzalez, ABL Larsen, JS~Vestergaard, AB~Dahl, et~al.,
\newblock ``Assessment of algorithms for mitosis detection in breast cancer
  histopathology images,''
\newblock {\em Med. Image Anal.}, vol. 20, no. 1, pp. 237--248, 2015.

\end{thebibliography}

\end{document}